\documentclass[letterpaper, 10 pt, conference]{ieeeconf}  

\IEEEoverridecommandlockouts                              

 \usepackage{subfigure}
\usepackage{graphics} 
\usepackage{amsmath} 
\usepackage{amssymb}  
\usepackage{siunitx}
\usepackage{mathtools}
\usepackage{booktabs}
\usepackage{multirow}
\usepackage[ruled]{algorithm2e}
\usepackage{pifont}
\newcommand{\cmark}{\ding{51}}%
\usepackage{siunitx}

\usepackage[dvipsnames]{xcolor}
\usepackage[font=small,labelfont=bf]{caption}
\usepackage{tikz}
\usepackage[bottom=57pt,top=54pt, left=54pt, right=54pt]{geometry}   

\usepackage{adjustbox}
\usepackage{pbox}
\usepackage{dblfloatfix}
\usepackage{graphicx}
\usepackage{hyperref}
\usepackage{caption}
\usepackage[nospace]{cite}

\usepackage{flushend}
\usepackage{tikz}
\usetikzlibrary{positioning}
\usepackage{pgfplots}
\usepackage{array,multirow,graphicx}
\usepackage{makecell}
\usepackage[T1]{fontenc}
\usepackage{microtype}
\usepackage{graphicx}
\usepackage{tikz}
\usetikzlibrary{positioning}
\usepackage{pgfplots}
\usepackage{xspace}
\usepackage[bottom]{footmisc}

\newcommand{\etal}{\emph{et~al.}}

\newcommand{\ie}{\emph{i.e.}}

\newcommand{\parag}[1]{\vspace{-4px} \vskip8pt \noindent \textbf{#1} \hspace{3px}}

\usepackage{soul}

 \newcommand{\removed}[1]{}
\newcommand{\OurMethod}{ICGNet} 

\title{\LARGE \bf
\OurMethod: A Unified Approach for Instance-Centric Grasping
}
\author{René Zurbrügg$^{1*}$, Yifan Liu$^1$, Francis Engelmann$^1$, Suryansh Kumar$^2$,\\ Marco Hutter$^1$, Vaishakh Patil$^1$,  Fisher Yu$^1$
\thanks{
\hspace{-20pt}
\begin{minipage}{1.0\columnwidth}
\vspace{-3pt}
This research was partially supported by the ETH AI Center, ETH Zürich Career Seed Award and RobotX grant.\ The authors are with $^1$ETH Zürich, $^2$Visual Computing, School of PVFA, Texas A\&M University. $^*$Corresponding author zrene@ethz.ch. 
\end{minipage}
}
}
\usepackage{bbding}
\usepackage{pifont}
\usepackage{wasysym}
\usepackage{amssymb}

\usepackage{hyperref}
\hypersetup{
    colorlinks=true,
    linkcolor=blue,
    filecolor=magenta,      
    urlcolor=cyan,
    pdftitle={ICG-Net: A Unified Approach for Instance Centric Grasping},
    pdfpagemode=FullScreen,
    }

\begin{document}

\maketitle
\thispagestyle{empty}
\pagestyle{empty}

\begin{abstract}
Accurate grasping is the key to several robotic tasks including assembly and household robotics.
Executing a successful grasp in a cluttered environment requires multiple levels of scene understanding:
First, the robot needs to analyze the geometric properties of individual objects to find feasible grasps.
These grasps need to be compliant with the local object geometry.
Second, for each proposed grasp, the robot needs to reason about the interactions with other objects in the scene.
Finally, the robot must compute a collision-free grasp trajectory while taking into account the geometry of the target object.
Most grasp detection algorithms directly predict grasp poses in a monolithic fashion, which does not capture the composability of the environment.
In this paper, we introduce an end-to-end architecture for object-centric grasping.
The method uses pointcloud data from a single arbitrary viewing direction as an input and generates an instance-centric representation for each partially observed object in the scene.
This representation is further used for object reconstruction and grasp detection in cluttered table-top scenes.
We show the effectiveness of the proposed method by extensively evaluating it against state-of-the-art methods on synthetic datasets, indicating superior performance for grasping and reconstruction.
Additionally, we demonstrate real-world applicability by decluttering scenes with varying numbers of objects. \\
Videos and Code \url{icgraspnet.github.io}.
\end{abstract}

\section{Introduction}
\label{sec:introduction}
The ability of robots to perform accurate and collision-free grasp maneuvers holds the potential for a wide array of applications in embodied intelligence\cite{embodied:intell:2015} such as assembly, pick and place, and packaging.
Despite impressive progress~\cite{mahler2017dex,Gualtieri2016HighPG,Breyer2021VolumetricGN,Sundermeyer2021ContactGraspNetE6,huang2022edge},
predicting grasps and their accurate pose from a single pointcloud remains a challenging task.
A successful grasp prediction requires not only understanding an object's geometry, but also its physical properties including its mass, shape and friction.
Beyond single object grasping this task is especially difficult in multi-object settings that exhibit clutter and strong occlusions limiting object visibility.

Current methods for grasping within cluttered environments primarily rely on pointcloud observations from a single viewpoint ~\cite{van2020learning,Jiang2021SynergiesBA, Mousavian_2019_ICCV, Breyer2021VolumetricGN,Sundermeyer2021ContactGraspNetE6, huang2022edge}. 
Generally, these methods directly operate on fully observed pointclouds to predict accurate, collision-free grasp poses, and achieve impressive results ~\cite{huang2022edge}.

However, existing work processes pointclouds on a holistic scene level, without explicitly reasoning about individual object instances. 
Applying these methods for pick-and-place or target-driven grasping tasks, often requires additional post-processing or external components.
For example, numerous methods rely on given (groundtruth) segmentation masks~\cite{ChavanDafle2021SimultaneousOR, li2021simultaneous,murali20206,yang2020deep}, object templates~\cite{danielczuk2019mechanical,lou2021collision} and predict object shapes by iteratively filtering each instance in the pointcloud~\cite{ChavanDafle2021SimultaneousOR}.
This introduces an extra level of complexity and often results in sub-optimal predictions, specifically under heavy occlusions.

Instead, our key contribution is to reason about grasping on an object level by explicitly modeling each individual instance which enables learning of shape priors and grasp affordances.
Interestingly, as our method allows to predict object shapes and instances,
it provides a clear interface for target-centric grasping
and directly supports collision checks for manipulated instances
which guarantees the collision-free removal and stable placements of unknown objects.

Specifically, we propose a unified architecture for instance centric grasp and shape prediction from single view pointclouds. 
The core idea of our method is to reason about an environment by extracting object embeddings on an instance level.
To this end, we introduce a sparse feature volume, consisting of volumetric- and surface features at multiple scales.
We then distill object-centric information into latent object embeddings through an iterative refinement process of masked cross- and self-attention. 
These features and object embeddings are used to model contact-based grasp affordances and object shapes as implicit fields. 
Additional object predictions such as semantics and pointwise instance assignments directly evolve from the refinement.

\begin{figure}[t]
\includegraphics[width=1.0\linewidth]{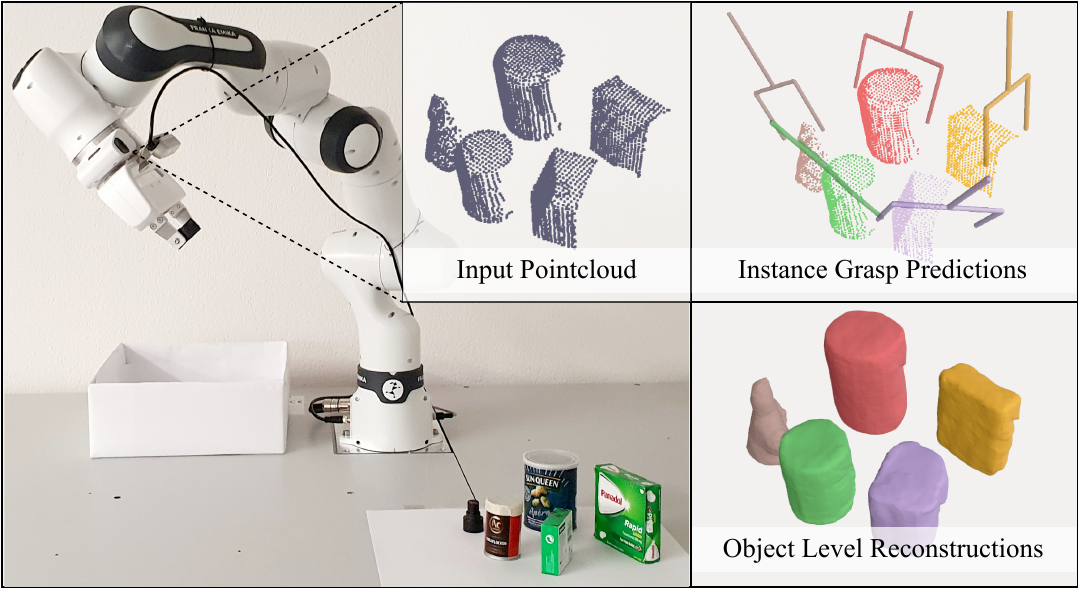}
\vspace{-15px}
\caption{\textbf{Overview of our network predictions}. Given a single view pointcloud we jointly predict instance segmentation masks, collision-free grasp predictions and reconstructions for each object. \label{fig:teaser}\vspace{-15px}}
\end{figure}


In experiments, significant improvements arise from our object-centric architecture,
establishing a new state-of-the-art in the \emph{packed} decluttering benchmark introduced in~\cite{Breyer2021VolumetricGN}
while surpassing scene-centric task baselines~\cite{Jiang2021SynergiesBA, huang2022edge}.
These improvements are reflected in real-world experiments deployed on the Franka Research 3 robot.
We additionally illustrate how instance-centric information can facilitate target-driven grasping (\ie{}, ``grasp instance number 1" or ``grasp the bottle") and effectively prevent post-grasp object-object collisions.

\section{Related Work}
\label{sec:rel_work}

\parag{Deep Grasp Synthesis.}
Recent advancements in robotic grasping have predominantly utilized deep learning techniques to detect robotic grasps directly from sensor data. Unlike more traditional heuristics~\cite{Newbury2022DeepLA}, these methods have demonstrated superior generalization performance towards previously unseen objects as well as achieving successful grasps in cluttered scenes.
While several methods have achieved high success rates in 4-degree-of-freedom (4\,DoF) top-down grasping~\cite{mahler2017dex,9926708}, their capabilities are often limited when dealing with cluttered scenes or task-dependent grasping scenarios~\cite{Mousavian_2019_ICCV}.
This limitation has led to an increased research focus on 6\,DoF grasp predictions, which aim to predict the full grasp pose from visual observations.
We further classify these methods in sampling- and scene-based approaches.
Sampling-based methods sample different grasp candidates and use a detector or diffusion \cite{weng2023neural, urain2023se} network to refine them further. GPD~\cite{Gualtieri2016HighPG} and PointNetGPD~\cite{liang2019pointnetgpd} learn to detect grasp poses in cluttered scenes from raw pointclouds by first sampling feasible grasp predictions and then using a CNN or PointNet classifier to predict the quality of each grasp based on the enclosed points.
6-DoF Graspnet~\cite{Mousavian_2019_ICCV} extends the way that grasp proposals are generated to the full SE(3) space by leveraging a variational auto-encoder for singulated objects.
Scene-based approaches directly predict feasible grasp poses for the whole scene in one forward pass.
Contact based methods such as Edge Grasp Networks~\cite{huang2022edge} or Contact GraspNet~\cite{Sundermeyer2021ContactGraspNetE6} directly predict the SE(3) pose, width, and grasp quality for each point in the pointcloud.
VGN~\cite{Breyer2021VolumetricGN} and GIGA~\cite{Jiang2021SynergiesBA} predict grasps for each voxel in the reconstructed TSDF using 3D CNN.

\parag{Simultaneous Shape Reconstruction and Grasp Estimation.}
Recently, joint prediction of scene reconstructions and grasp poses have been studied in more detail~\cite{Jiang2021SynergiesBA, varley2017shape, van2020learning, ChavanDafle2021SimultaneousOR} as both tasks are correlated and fundamental for environment interactions.
Early works by Varley \etal{}~\cite{varley2017shape} voxelized an observed pointcloud and utilized a 3D CNN as well as a marching cube algorithm for reconstruction.
The reconstructed mesh is then used in combination with GraspIt!~\cite{miller2004graspit} to predict feasible grasp candidates.
PointSDF~\cite{van2020learning} directly learns a signed distance field from the initial pointcloud and conditions a grasp classification on the latent representation of the occupancy decoder.
ShellGrasp-Net~\cite{ChavanDafle2021SimultaneousOR} jointly learns the camera-ray intersections with singulated objects by predicting grasp-affordances as well as entry and exit-depth maps.
GIGA~\cite{Jiang2021SynergiesBA} extends grasp-prediction and shape reconstruction to cluttered scenes.
Their methods combine VGN~\cite{Breyer2021VolumetricGN} and Convolutional Occupancy Networks~\cite{peng2020convolutional} to learn grasps and occupancy as continuous functions over 3D coordinates. They show that learning both tasks jointly can improve the grasp detection and reconstruction of graspable regions.

\parag{Target Driven Grasping.}
Prior work on grasp prediction typically deals with singulated objects~\cite{varley2017shape, ChavanDafle2021SimultaneousOR,jeng2021gdn} or predicts grasp affordances for the full scene without any notion of instances~\cite{Breyer2021VolumetricGN, Jiang2021SynergiesBA, liang2019pointnetgpd}.
While these methods can be adapted for target-driven grasping by segmenting relevant objects and predicting grasp poses for them, the resulting grasps may not guarantee collision-free execution.
To address this, Murali \etal{}~\cite{murali20206} introduce a collision network that post-processes predicted grasps and discards or refines those that would cause collisions.
\cite{breyer2022closed} rely on accurate bounding boxes of the objects to post-process scene-based grasps,
assuming that the fingertips of the gripper lie inside the bounding box of the target object.
Sundermeyer \etal{}~\cite{Sundermeyer2021ContactGraspNetE6} propose contact-based grasp detectors that directly predict grasp poses for each observable point in the input pointcloud of a cluttered scene.
This enables the selective grasping of objects based on the semantic class of the associated contact point.
However, all these approaches rely on external segmentation modules to predict instance segmentation masks, adding complexity to the overall pipeline. Contrary to that, our work introduces a unified method for scene understanding which combines panoptic segmentation, reconstruction, and grasp detection.

\section{Method}

\begin{figure}[t!]
    \vspace{-5px}
\centering
\includegraphics[width=.7\linewidth]{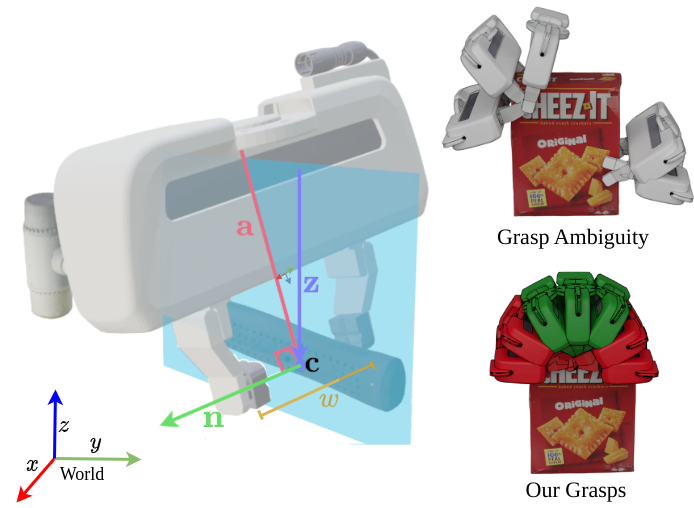}

\caption{\textbf{Grasp Representation.} \emph{Left:}
    $\mathbf{c}$ is the contact point of the closed gripper and
    $\mathbf{n}$ is its estimated surface normal. $\mathbf a$ is the approach direction of the gripper. Given the gravity vector $\mathbf z$ and surface normal $\mathbf n$, $\mathbf a$ can be uniquely defined by the approach angle $\alpha$.
\emph{Top Right:} Grasp ambiguity of different grasp representations. When dealing with a particular contact or gripper center, there can be multiple feasible approach direction resulting in a successful grasp. 
\emph{Bottom Right:} For each contact point, our representation enables the prediction of grasp qualities for different gripper orientations perpendicular to the surface normal. \vspace{-10px}}
    \label{fig:grasp-repr}
\end{figure}
\begin{figure*}[t!]
    \centering
    \includegraphics[width=0.87\linewidth]{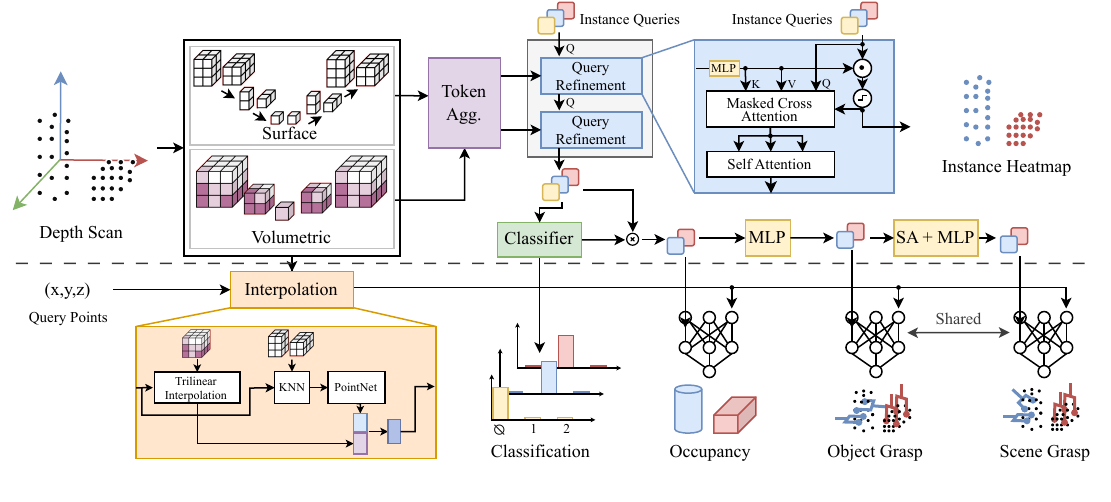}
\vspace{-15px}
    \caption{\textbf{Model Overview.} Given an input pointcloud, we voxelize the pointcloud and extract volumetric and surface features at multiple scales using a sparse Minkowski-~\cite{choy20194d} and dense U-Net~\cite{cciccek20163d}.
    The surface features are enriched with volumetric information and treated as tokens with positional encodings based on voxel locations.
    Masked attention iteratively refines instance queries by cross-attending to extracted sparse tokens.
    This process allows each latent query to focus on a specific instance and to be classified as \emph{``<semantic class>''} or \emph{``no object''}.
    The refined queries condition the task-specific decoders to model the occupancy of each instance directly or to predict grasp affordance scores and gripper widths.}
    \label{fig:overview}
\vspace{-15px}
\end{figure*}
\begin{figure}[t]

\vspace{-0px}
\centering
\small
\includegraphics[width=0.38\linewidth, trim={0 0 6 0},clip]{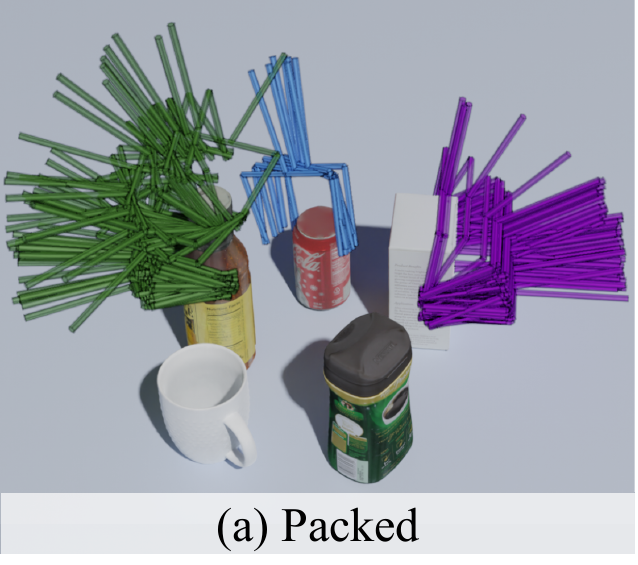}
\hspace{3mm}
\includegraphics[width=0.38\linewidth, trim={0 00 0 0},clip]{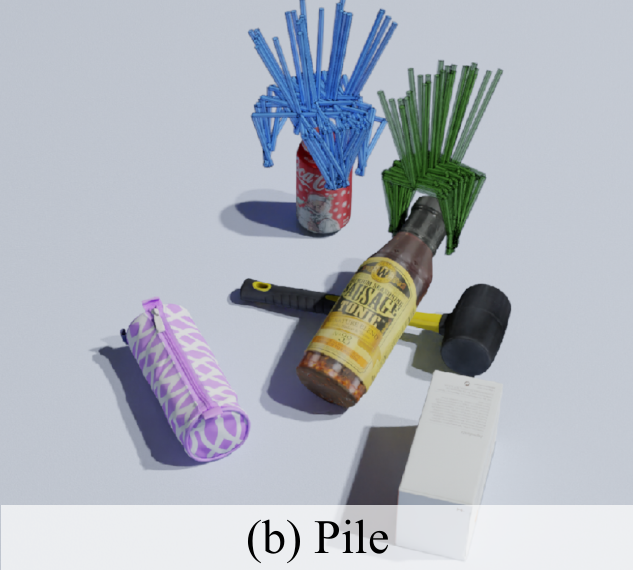}\\

\caption{
\label{fig:instance-pred-sim}
Our grasp predictions on simulated,
unseen test objects from\cite{downs2022google}.
Predicted grasps for ``\textcolor{OliveGreen}{bottle}'', ``\textcolor{RoyalBlue}{can}'' and ``\textcolor{violet}{box}'' in the packed \emph{(left)} and pile setup \emph{(right)}. More qualitative examples at can be found on the project page}
\vspace{-15px}
\end{figure}

\vspace{-2px}
\parag{Problem Formulation.}
Our setup consists of a robot arm with a parallel-jaw gripper operating within a planar tabletop workspace as shown in Fig.~\ref{fig:teaser}.
The workspace contains multiple rigid objects placed on the tabletop. The objects are either placed randomly in an upright position (packed) or dumped from a box (pile) (Fig.~\ref{fig:instance-pred-sim}).
Prior to each object interaction, the scene is captured once using a depth camera from a static, randomized position following~\cite{huang2022edge}.
The captured depth image is converted into a pointcloud and fed into the proposed model. The model jointly reconstructs the full 3D shape of each object and predicts stable\footnote{Grasps which are able to steadily hold the object even under minor movement or perturbations.} grasps that do not collide with the other objects and the environment.

Formally, given a pointcloud $P_{scan}\in \mathbb{R}^{N_s\times3}$ consisting of $K$ different objects from $C$ different classes,
we predict per-point instance labels $I \in \{1,\dots,K\}^{N_p}$ and semantic labels $S \in  \{1,\dots,C\}^{N_p}$.
To reconstruct the object shapes, we additionally predict occupancy values $o \in \{0,1\}^{N_q}$ for a set of $N_q$ query points $P_{query} \in \mathbb{R}^{N_q\times 3}$ for each instance. Finally, we predict grasp affordances (\ie{} the probability of a grasp being successful) for $N_\alpha$ different approach directions $\mathcal{A} \in [0,1]^{N_{\alpha} \times N_q}$ for each instance.

\parag{Grasp Representation.}
\label{sec:grasp-repr}
Existing grasp prediction methods typically rely on a single ground truth grasp for a given contact point~\cite{Sundermeyer2021ContactGraspNetE6} or gripper center position~\cite{Breyer2021VolumetricGN, ChavanDafle2021SimultaneousOR}. However, when a fixed contact point on an object is considered, there can be multiple equally "good" grasp orientations for removing the object from the scene (Fig.~\ref{fig:grasp-repr}, \emph{left}). In line with~\cite{huang2022edge}, we argue that the \emph{distribution} of valid grasps given a query point provides more consistent supervision. This approach aims to reduce the number of positive predictions that are mistakenly treated as "wrong" while generating more diverse grasp proposals.
Nonetheless, learning a continuous distribution over all potential grasp poses for each contact point is intractable. To address this, we enforce the approach direction $\mathbf{a}$ to be perpendicular to the surface normal, thereby restricting the approach vector to lie on a plane given a contact point\footnote{While this limits the diversity of grasps, the assumption is often closely met in practice when requiring force closure of the resulting grasp.}.
Furthermore, we discretize the approach direction into a set of discrete angles and formulate the grasp prediction as a multi-class classification problem (Fig.~\ref{fig:grasp-repr}, \emph{bottom-right}). With this, our contact-based grasps take the following form:
\begin{equation}
    \mathcal{G}_{\text{contact}} = (\mathbf{c},\mathbf{n},\mathbf{s},w) ,
\end{equation}
$\mathbf{c} \in \mathbb{R}^3, \mathbf{n} \in \mathbb{R}^3, \mathbf{s} \in [0,1]^{N_{\alpha}}, w \in [0, w_{max}]$ with $\mathbf{c}$ being the contact point, $\mathbf{n}$ the surface normal, $\mathbf{s}$ the grasp affordance (= probability of successful grasp) values for each discretized approach direction and $w_{max}$ being the maximal opening width of the gripper. 

The corresponding set of SE(3) poses for a given grasp affordance prediction $g$ and gravity vector $\mathbf{z}$ is given by the tool-center point and computed as 
\begin{equation}
    \mathcal{G}_{SE3} = \{(\mathbf{t}_g, R_g^i)\}_{i=0}^{N_\alpha-1} \text{ with }
   \mathbf{t}_g = \mathbf{c} + \frac{w_{max} - w}{2} \mathbf{n},
\end{equation}
\begin{equation}  
R_g^i = R_y(\alpha_i) \cdot \begin{bmatrix}
    | & | & | \\
     \mathbf{z} \times \mathbf{n} & \mathbf{n}  & ( \mathbf{z} \times \mathbf{n} ) \times \mathbf{n} \\
    | & | & | \\
\end{bmatrix},
\end{equation}

where the approach angle $\alpha_i$ is within $\{-90^\circ, \dots, 90^\circ\}$ and $R_y$ denotes to rotation matrix around the y axis.

\emph{Singularity}: The proposed grasp representation results in a singularity when the surface normal and gravity vector coincide. If this is the case, (|$\mathbf{z} \cdot \mathbf{n}| > 0.98$), the x-axis of the grasp is chosen to align with the table surface pointing in the arbitrary x direction of the world frame.

\parag{Model Architecture.}
\label{sec:model-arch}
Fig.~\ref{fig:overview} shows our end-to-end instance aware grasp prediction model. It consists of two stages: 

\noindent\textbf{Encoder.}
Given a pointcloud in the world frame, our encoder network performs several key tasks. It extracts both sparse and dense features at multiple resolutions utilizing a sparse 3D-UNet architecture~\cite{choy20194d} for surface features and dense 3D-UNet architecture~\cite{cciccek20163d} for volumetric features. The sparse features are enriched with volumetric information through the proposed \emph{token aggregation} module. To decompose the voxelized scene into individual instances, we apply multiple \emph{instance query refinement} modules similar to Mask3D~\cite{schult2022mask3d}.
Further, we rely on a \emph{classification head} to assign a class to each latent representation and filter out unmatched queries using a \emph{no-object} class. All valid instance queries are directly used as latent embeddings for an occupancy network. Additionally, we employ a final Self Attention and MLP layer to exchange inter-object information between the queries and convert them to the affordance domain.

\noindent\textbf{Decoder.}
Our decoder is designed as an implicit neural field and uses world coordinates ($x,y,z$) as input. It predicts occupancy $p_{occ}$ and grasp affordances $g \in \mathcal{G}_{aff}$ for each instance.
In addition, predictions such as classification and instance heatmaps are directly extracted from the encoder as shown in Fig.~\ref{fig:overview}. These predictions are made for each instance query (class) and each point within the input pointcloud (instance id). To model each implicit decoder, we use a series of MLPs with residual connections following the approach proposed in~\cite{Mescheder2018OccupancyNL}. Further, we concatenate each instance query with its positional encoding, which captures spatial information.
Moreover, the queried coordinates ($x,y,z$) are enriched by incorporating the surface and volumetric features using our \emph{interpolation} module. These enriched coordinates are then concatenated with the positional encodings of the original coordinates.

\noindent\textbf{Token Aggregation.}
Ideally, our goal is to extract dense features at a high resolution and directly feed them into the instance query refinement module. However, this approach comes with significant drawbacks, notably increased memory consumption and computation cost, rendering it infeasible for most applications. On the other hand, sparse neural networks scale with the number of occupied voxels, making them suitable for scaling to large areas. This advantage arises as the number of occupied voxels typically grows slower than the total number of voxels. Here, we combine the best of both by enriching each occupied surface feature with a volumetric context that is extracted from the volumetric feature grid. To this end, each surface feature is concatenated with the feature of the nearest volumetric voxel, similar to PointNet~\cite{qi2017pointnet}. This enables the volumetric feature backbone to operate on a larger resolution extracting context information to enrich the sparse features used in the query refinement.

\noindent\textbf{Query Refinement.}
Given a set of $K$ instance queries, we apply a series of masked cross-attention and self-attention to extract instance-centric information given the scene-level features extracted from the U-Net backbones.
We adapt Mask3D~\cite{schult2022mask3d} and add an MLP layer to further process the extracted scene features.
We also add Fourier positional encodings~\cite{tancik2020fourier} based on voxel positions and use farthest-point sampling to sample initial instance query positions. 

\begin{table}[t]
    \centering
    \setlength{\tabcolsep}{5pt}
    \begin{tabular}{l|rrr|c|c|c}
    \toprule
    Method & G & R & S & \#Params & Inference & Latency \\ \midrule
        VGN~\cite{Breyer2021VolumetricGN} & \cmark & -- & -- & 0.3 M  & 3\,ms & 7\,ms \\
        GIGA~\cite{Jiang2021SynergiesBA} & \cmark & \cmark & --& 0.6 M & 26\,ms& 30\,ms \\
        GIGA-HR~\cite{Jiang2021SynergiesBA} & \cmark & \cmark & --& 0.6 M & 56\,ms& 66\,ms \\
        EdgeGraspNet~\cite{huang2022edge} & \cmark & -- & --& 3.0 M & 26\,ms& 34\,ms \\
        VN-EdgeGraspNet~\cite{huang2022edge} & \cmark & -- & --& 1.7 M & 264\,ms& 306\,ms \\
        \OurMethod{} (Ours) & \cmark & \cmark & \cmark & 3.2 M & 137\,ms & 138\,ms\\ 
    \bottomrule
    \end{tabular}
    \caption{\textbf{Tasks, Model Size, and Runtime.} We report the supported tasks, number of parameters, model inference times, and average latency (including grasp postprocessing) measured on a GeForce RTX 3080. Although, our network consists of slightly more parameters compared to~\cite{huang2022edge}, ICGNet (Ours) is the only architecture that is able to predict grasps (G), reconstruction (R) and semantics (S) while having a lower latency than VN-EdgeGraspNet~\cite{huang2022edge}.}
    \label{tab:num_paras}
\vspace{-15px}
\end{table}
\noindent\textbf{Classifier.} 
The classifier predicts a class label including a \emph{non-object} class to address the varying number of instances in a scene.
A small MLP followed by a softmax activation predicts a categorical distribution over the desired $C + 1$ classes.
For our experiments, we manually annotate the points with class labels that correlate with the respective shape, consisting of six categories
(``mug", ``box", ``can", ``bottle", ``cylindric", ``ball" and ``other").
Having the notion of semantic classes, allows to predict grasp candidates and reconstructions for individual objects as shown in Fig.~\ref{fig:instance-pred-sim}.

\noindent\textbf{Interpolation.}
Directly passing the input coordinates or the respective positional encoding to the occupancy network produces sub-optimal reconstructions and grasp predictions due to the limited scene information.
Therefore, the input coordinates are concatenated with per-point features extracted from the sparse and dense feature grids.
Volumetric features are extracted using trilinear interpolation.
Interpolated features from the sparse surface volume stem from a K-Nearest Neighbour (KNN) search. Naively averaging the KNN leads to identical features for points on opposite sides of the surface.
We therefore use a small PointNet for feature aggregations which relies on the KNN features and distance to the nearest neighbors.
Both dense and sparse features are concatenated and fed through a MLP to extract the enriched embeddings for each coordinate. 

\parag{Loss Formulation.}
The proposed model is trained in an end-to-end fashion. The training uses instance and semantic annotations as well as occupancy and grasp poses obtained from simulation.
The implemented loss consists of binary cross-entropy, DICE and squared-error losses and is given as
\begin{equation}
\begin{split}
    \mathcal{L} &=  \mathcal{L}_\mathtt{inst}^{BCE} +  \mathcal{L}_\mathtt{inst}^{DICE} + 
    \mathcal{L}_\mathtt{sem}^{BCE} + \\ 
  &\ \ \ \ 
    \mathcal{L}^{Grasp} + \mathcal{L}_\mathtt{occ}^{BCE},
\label{eq:full-loss}
\end{split}
\end{equation} 
with\vspace{-4mm}
\begin{equation}
    \mathcal{L}^{Grasp} = \mathcal{L}^{BCE} + \left\|\cdot \right\|_2^2,
\end{equation}
where the first three terms refer to the panoptic segmentation task and are calculated with respect to the input pointcloud $P_\mathtt{scan}$. $\mathcal{L}^{Grasp}$ refers to the grasp predictions for each instance. This loss is calculated with respect to the sampled grasp coordinates which may differ from the actual observed pointcloud. We use the cross entropy loss to supervise the approach classification and the squared L2 loss for the gripper width. Finally, the reconstruction loss $\mathcal{L}_\mathtt{occ}^{BCE}$ supervises the predicted occupancy for each instance and is computed with respect to an additional set of randomly sampled points within the scene. More information about the labels and dataset is provided in Sec. \ref{dataset}.
Note that there is no ordering in the set of instances in a scene and we need to establish correspondence between the set of predicted and set of groundtruth instances during training.
Following~\cite{schult2022mask3d}, we use Hungarian matching based on the instance segmentation\footnote{We limit the matching to the segmentation task since calculating every possible assignment of the occupancy field is computationally extensive.} loss to find unique assignments and apply the same loss function to the predictions at multiple resolutions.

\begin{table*}[t!]
    \setlength{\tabcolsep}{6pt}
    \centering
    \begin{tabular}{lrr|cc||cc|cc}
    \toprule
         &  \multicolumn{4}{c||}{Grasping} & \multicolumn{4}{c}{Reconstruction} \\
         &  \multicolumn{2}{c}{Packed}& \multicolumn{2}{c||}{Pile}&\multicolumn{2}{c}{Packed}& \multicolumn{2}{c}{Pile} 
        \\ Method &  GSR (\%) $\uparrow$ & DR (\%) $\uparrow$  &GSR (\%) $\uparrow$ & DR (\%) $\uparrow$ & C-L1 [mm] $\downarrow$ & IoU (\%) $\uparrow$ & C-L1 [mm] $\downarrow$ & IoU (\%) $\uparrow$ \\ \midrule
        PointNetGPD ~\cite{liang2019pointnetgpd}&  79.3 $\pm$ 1.8 & 82.5  $\pm$  2.9 &  75.6  $\pm$  2.3 & 77.0 $\pm$  2.8 & -- & --  & -- & --\\
        VGN~\cite{Breyer2021VolumetricGN} &   80.2  $\pm$  1.6 & 86.2  $\pm$  2.0  & 64.9  $\pm$  2.2 & 69.1  $\pm$  3.2  &-- & -- & -- & -- \\
        GIGA$^\dagger$~\cite{Jiang2021SynergiesBA} & 89.9 $\pm$ 1.7 & 87.6 $\pm$ 2.0  &76.3 $\pm$ 2.4 & 80.9 $\pm$ 4.1  &  2.5 $\pm$ 1.2  & {82} $\pm$ 6.9 &  3.4 $\pm$ 1.4 &  71 $\pm$ 8.6   \\  
        GIGA-HR$^\dagger$~\cite{Jiang2021SynergiesBA} & \underline{91.4} $\pm$ 1.5 & 88.5 $\pm$ 1.4  & 86.5 $\pm$ 1.2& 80.8 $\pm$ 1.9 &  2.5 $\pm$ 1.2  & {82} $\pm$ 6.9 &  3.4 $\pm$ 1.4 &  71 $\pm$  8.6    \\  
EdgeGraspNet$^*$\cite{huang2022edge} & 90.6 $\pm$ 0.9 & \underline{93.9} $\pm$ 0.7 & 91.0 $\pm$ 2.0 & \underline{93.7} $\pm$ \underline{2.3}& -- & --  & -- & -- \\ 

 VN-EdgeGraspNet$^*$\cite{huang2022edge} & 90.4 $\pm$ 2.5 & 92.8 $\pm$ 1.0  & \underline{91.9} $\pm$ \underline{0.8} & 92.7 $\pm$ {1.2} & -- & --  & -- & -- \\

 \OurMethod{} (Ours) & \textbf{97.7}	$\pm$ \textbf{0.9} & \textbf{97.5} $\pm$ \textbf{0.3} & \textbf{92.0} $\pm$ \textbf{2.6}
 & \textbf{94.1} $\pm$ \textbf{1.4}&  \textbf{2.3} $\pm$ \textbf{1.1} & \textbf{84} $\pm$ \textbf{6.1}&  \textbf{2.9} $\pm$ \textbf{1.9} & \textbf{77} $\pm$ \textbf{9.9}  \\

\bottomrule
\end{tabular}
\caption{\textbf{Comparison to State-of-the-Art on Synthetic data.}
    Simulated results on the packed and piled scenes using the evaluation setup from \cite{huang2022edge}. We report grasp success rate (GSR), declutter rate (DR), and reconstruction performance, Chamfer L1 distance (C-L1), IoU. \OurMethod{} denotes our approach that predicts grasp poses and occupancy for each individual instance. We retrain ($^\dagger$) GIGA on randomized viewpoints and 2M grasps for each environment separately and re-evaluate ($^*$) the pre-trained checkpoints provided by \cite{huang2022edge} multiple times. The scores for VGN and PointnetGPD are taken from \cite{huang2022edge}. Highest number marked in bold, second highest underlined. }
    \label{tab:giga_sim_res}
\vspace{-15px}
\end{table*}
 
\vspace{-1mm}
\section{Experiments}
\label{sec:experiments}



\parag{Training Details.}
Our model is trained on simulated data and makes use of zero-shot to transfer to a real Franka Research 3 arm from Franka Emika
We train the proposed network for 60 epochs with an effective batch size of 8 using AdamW~\cite{loshchilov2017decoupled} with a learning rate of $1\mathrm{e}{-3}$ and a linear warmup, cosine annealing learning rate schedule~\cite{loshchilov2016sgdr}. Additionally, we make use of early stopping based on the F1 score of the grasp affordances calculated on the validation set. We find the F1 score to be a more robust performance metric as the instance-wise affordance scores are heavily imbalanced. The full training takes $\sim$45h on a Nvidia Titan RTX GPU.

\parag{Simulation Environment and Dataset.}
\label{dataset}
We leverage the simulation setup and the object dataset introduced in VGN~\cite{Breyer2021VolumetricGN}. This dataset consists of 303 train and 40 test objects from various sources~\cite{calli2017yale,kit_ds,Singh2014BigBIRDAL}. The experimental setup involves a free-floating Franka Emika Gripper and we sample grasps and occupancy values in $30$ cm$^3$ sized tabletop workspace. 
Contrary to ~\cite{Breyer2021VolumetricGN, Jiang2021SynergiesBA}, which randomly samples the gripper centers close to the surface and stores the resulting best gripper orientation as a ground truth label, we adapt a different strategy. We select an observed point and corresponding surface normal, then execute the grasp from different approach angles as outlined in Sec.~\ref{sec:grasp-repr}. Specifically, we sample twelve different approach angles for each contact point and the final gripper width as well as the grasped object are recorded. 
Additionally, we sample the occupancy values for each scene. A total of 200'000 occupancy values are sampled of which 70\% are uniformly sampled and 30\% are sampled closer to the object surfaces to allow more accurate reconstruction. We store the occupancy value for each object yielding a total of 200'000 $\times$ $k$ binary labels for each scene.
We further make use of the \emph{packed} and \emph{piled} splits introduced by~\cite{Breyer2021VolumetricGN} and sample a total of 1M and 2M grasps from each split, resulting in 3M grasps from 15'000 scenes of which 13'500 and 1'500 are used for training and validation respectively.

\parag{Observations.}
For each scene, a single depth-image is captured and the unprojected pointcloud in the world frame is used as the input to our network.
We randomize the depth camera pose to be uniformly located at the spherical coordinates ($r\in [0.48, 0.72], \theta \in [0, \frac{\pi}{4}], \phi \in [0, 2 \pi])$ looking at the center of the workspace.
To allow for better sim-to-real transfer, we further randomly rotate the pointcloud, add noise sampled from $\mathcal{N}(0, 0.002)$ and apply 3D elastic deformation~\cite{vantulder2021}.
In addition, the scanned pointcloud is divided into regions of $2$\,cm, which are randomly erased with probability $p=0.2$. The tabletop is removed based on the points height.

\parag{Grasp Selection and Reconstruction.}
For each scene, 32 instance queries are spawned and refined using our query refinement and the observed pointcloud. The pointcloud is denoised using statistical outlier filters. It is then downsampled using a voxel size of $2$\,mm before computing the grasp affordances for each point. Surface normals for each contact point are estimated using covariance analysis from Open3D~\cite{zhou2018open3d}.
We then execute the grasp associated with the highest affordance score which is collision-free, given our implicit shape encodings of the scene and height of the table. 
To evaluate the scene reconstruction quality, we compute the occupancy field for the whole scene by combining instance level predictions $p^{scene}_\mathtt{occ}(x) = \max_{\mathcal{I} \in  Inst} p_\mathtt{occ}(x | \mathcal{I})$ and use adaptive marching cubes algorithm~\cite{Mescheder2018OccupancyNL}.

\parag{Simulation Results.}
We evaluate our method using the evaluation process from~\cite{huang2022edge} for the \emph{pile} and \emph{packed} scenes.
For the \emph{grasping} task, we report the mean and standard deviation of over 4 different runs, each consisting of 100 different scenes. We report the Grasp success rate (GSR, percentage of successful grasps) and declutter rate (DR, percentage of removed objects after the task has finished) 
For the \emph{reconstruction} task, we randomly sample 100 different \emph{pile} and \emph{packed} scenes using the objects from the test split.
The Chamfer L1 and IoU scores are calculated following~\cite{Mescheder2018OccupancyNL}. 
Scores are shown in Tab.~\ref{tab:giga_sim_res} where
\OurMethod{} exhibits superior or equal performance compared to other methods in both grasp success and reconstruction, achieving the highest GSR, DR, IoU and lowest Chamfer L1 distance. Our method significantly outperforms the baseline methods on the \emph{packed} dataset, highlighting the advantage of our grasp representation and instance priors. We additionally compare the model size and performance scores in~\ref{tab:num_paras}, showing that our model is competitive in size and inference speed while combining multiple tasks.

\parag{Challenging Viewpoints.}
We further evaluate our method on the challenging task of grasp detection from top-down viewpoints. To this end, we modify the evaluation algorithm from~\cite{huang2022edge} to spawn the camera facing downwards at a random angle $\theta \in [0,2^\circ]$. This setup is extremely challenging for contact-based grasping since almost no graspable regions are observed and the surface normals are harder to estimate, resulting in significantly lower declutter rates of (85\% and 72\%) on the packed and pile environments respectively. To overcome this limitation of contact based grasping, we leverage our multi-task architecture and resample, unobserved points from the (implicit) object surfaces if no feasible grasp is found. We then use these new surface points for grasp prediction, allowing us to improve the declutter rate to \textbf{90}\%(\emph{+6\%}) and \textbf{94}\% (\emph{+31\%}) respectively.

\begin{figure}[t!]
    \centering
\begin{tikzpicture}
    \footnotesize
    \node at (0, 0) {\includegraphics[width=.35\textwidth, trim={0 50 50 110},clip]{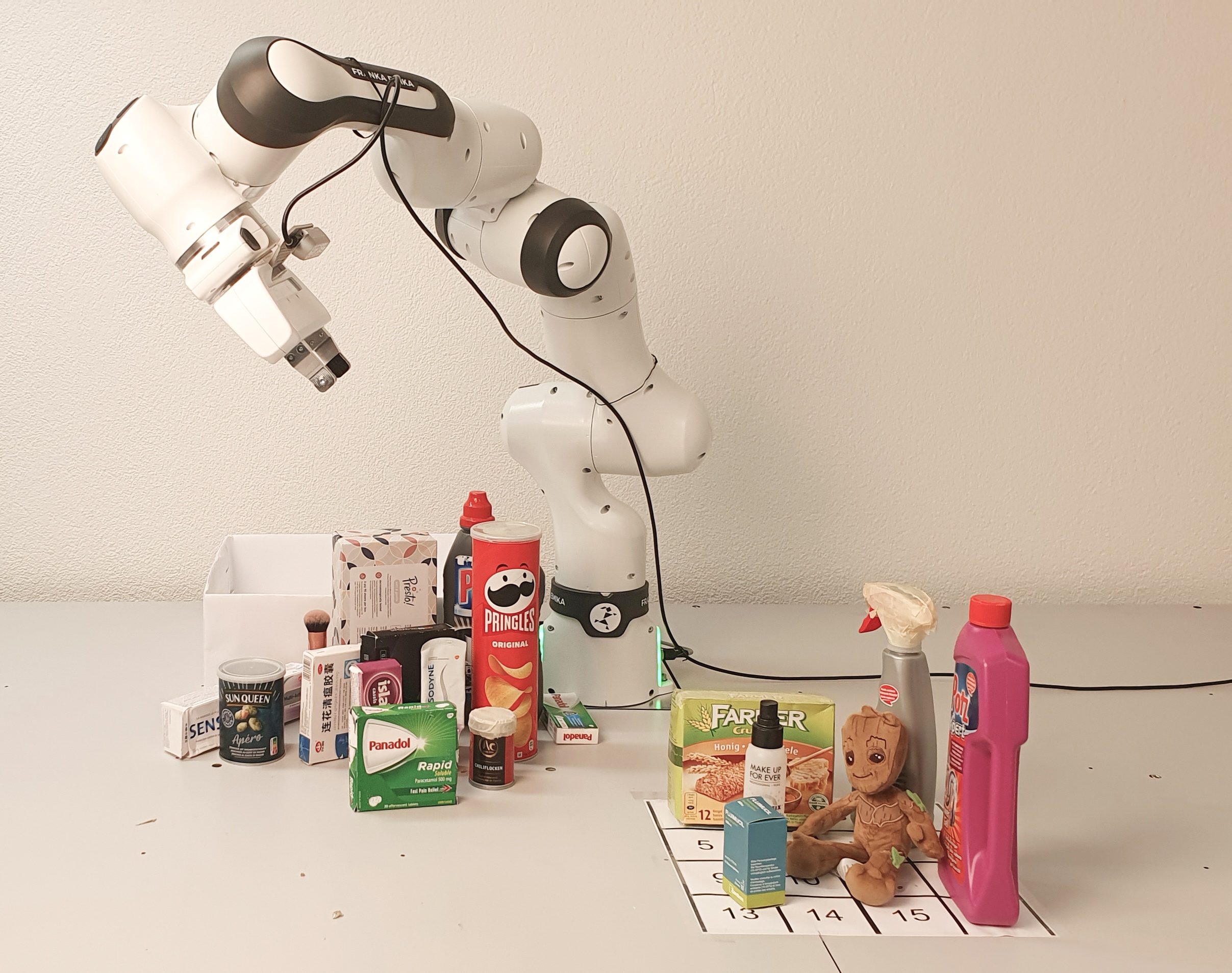}};
    \node[anchor=east] (cam) at (2.4, 1.9) {Depth Camera};
    \draw[red, thick] (cam) -- (-1.7, 1.35);
    \node[anchor=east] (box) at (-1.9, .2)  {\parbox{\widthof{Drop-Off}}{Drop-Off\\Box}};
    \draw[red, thick] (box) -- (-1.9, -.3);
    \node[anchor=east] (obj) at (2.45, .3)  {\parbox{\widthof{Objects}}{Test\\Objects}};
    \draw[red, thick] (obj) -- (-0.45, -.7);
    \draw[red, thick] (obj) -- (1.7, -1.3);
    \node[anchor=east] (benchmark) at (-0.6, -2.0)  {Grid (Repeatability)};
    \draw[red, thick] (benchmark) -- (0.35, -2.0);
\end{tikzpicture}
\caption{
\textbf{Real world experimental setup.}
We use 17 different objects of which 3-6 are placed on a $30$\,cm$^3$ workspace. }
\label{fig:experiment-setup}
\vspace{-15px}
\end{figure}
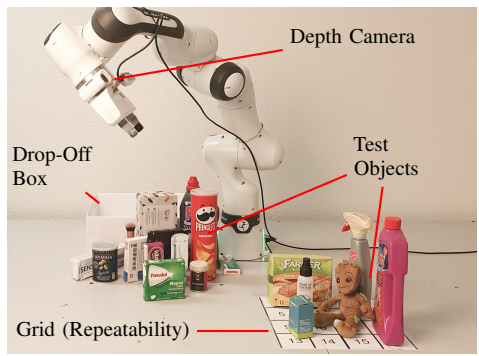

\parag{Real World Experiments.}
We validate our method and compare it against GIGA~\cite{Jiang2021SynergiesBA} and VN-EdgeGraspNet~\cite{huang2022edge} on different real-world declutter tasks. The experimental setup is depicted in Fig.~\ref{fig:experiment-setup}. A RealSense D415 RGB-D camera is attached to the gripper on a Franka Research 3 Arm. 
We use a $30 \times 30$\,cm workspace to allow for comparison with ~\cite{Jiang2021SynergiesBA}.\footnote{Note that our method is capable of inferring grasps on larger scenes due to the combination of dense and sparse voxelization.}
We use a total of 17 different test objects. 
Of these, 3-6 objects are randomly placed on a $30 \times 30$\,cm workspace.
Before each experiment, we collect a top-down image of all objects, which are placed on a grid to ensure the repeatability and a fair comparison of all methods.
Before each grasp trial, the arm moves to the designated image acquisition position (see Fig.~\ref{fig:experiment-setup}) and captures depth image, which is post-processed and fed into the network. We use MoveIt!~\cite{coleman2014reducing} to plan and execute collision-free grasp maneuvers. For collision checking, we rely on the predicted scene reconstructions or use the collision mesh from the tsdf volume.
%
Given a set of predicted grasp poses, we follow the procedure of \cite{huang2022edge} and filter the grasps for a 0.9 confidence level and execute the pose with the highest z component that is kinematically feasible and collision-free. If no grasp is found, we lower the confidence level to 0.8 and 0.7. The current task is terminated if two consecutive grasp failures occur or if no grasps are found for five sequential observations. Grasp success rate (GSR) and declutter rate (DR) for different clutter categories are reported in Tab.~\ref{tab:real-world-res}.

\begin{table}[]

\centering
\resizebox{\columnwidth}{!}{

\begin{tabular}{l|cc|cc}
\toprule
Method &  \multicolumn{2}{c|}{Packed}& \multicolumn{2}{c}{Pile}
    \\  &  GSR $\uparrow$ & DR$\uparrow$ & GSR (\%) $\uparrow$ & DR (\%) $\uparrow$ \\ \midrule
    GIGA-HR$^{[1]}$ &   73\% & 59\%  & \underline{83}\%  & \underline{71}\% \\
    VN-EdgeGraspNet$^{[2]}$ & \underline{84}\% & \underline{76}\%  & 81\% & \underline{71}\% \\
    \OurMethod{}  (Ours) &   \textbf{90}\% & \textbf{88}\%  & \textbf{90}\% & \textbf{83}\%\\
    \bottomrule

\end{tabular}

}
\caption{\textbf{Clutter Removal Experiment on Real-World Data.} We report the grasp success rate (GSR) as well as declutter rate (DR) evaluated in the real world setting.\vspace{-15px}} 

\label{tab:real-world-res}
\end{table}


We observe that the reported success and declutter rates follow the same trends but are lower than what is usually reported in~\cite{huang2022edge,Jiang2021SynergiesBA} (grasp success rates of $90\%$ and almost perfect declutter rate). We attribute this to our hardware setup (a panda gripper that has a bigger collision shape and smaller gripper width than the robotiq gripper used in~\cite{Jiang2021SynergiesBA}) and the object set, which is quite challenging as some materials are slightly reflecting. Additionally, our test objects often require the gripper to be fully open and some objects might not be re-grasped once tipped over due to their height or aspect ratio, prohibiting the scene from being fully decluttered. Our results show that our network successfully transfers to the real world and achieves higher success and declutter rates compared to~\cite{huang2022edge,Jiang2021SynergiesBA}. 
Additionally, we observe an increased amount of object-object collisions for~\cite{huang2022edge,Jiang2021SynergiesBA} when moving to the drop-off location, as their method does not directly support collision checks with the grasped objects. Our method allows for instance-aware reconstructions of the scenes and can therefore anticipate the attached object geometry and often finds a collision-free path to the drop-off location. 

\parag{Limitations.}
Although our architecture demonstrates remarkable performance, we have identified certain limitations. In line with~\cite{schult2022mask3d}, we observe that our network occasionally over-segments or combines instances even when they are not in contact. Most of these errors can be corrected by post-processing the network predictions, which comes at the cost of higher latency. Additionally, we observe that the reconstructions can be further improved by learning occupancy for the full scene instead of for each instance independently.    
We plan to address these limitations in future work by utilizing dilated attention to 
to improve spatial coverage of the query and include color into the pointcloud features. Additionally, we plan to train our network on datasets with diverse objects in order to learn robust object priors. 
\section{Conclusion}
\label{sec:conclusion}
We introduce \OurMethod, a unified architecture for target-driven grasping in cluttered environments that allow for instance centric, target-driven grasping and object reconstruction from a single view pointcloud. 
We evaluate our network performance both in simulation and challenging real-world scenes. Our results show, that the proposed method outperforms the current state of the art for grasping in cluttered environments and can be successfully transferred to the real world. 
A clear direction for future work includes learning more robust instance priors on more extensive and  diverse datasets and including language embeddings for target-driven grasping. 
\clearpage

{\small
\bibliographystyle{IEEEtran}
\bibliography{IEEEfull,root}
}

 \end{document}